\documentclass{article}

\usepackage[round]{natbib}
\usepackage{fancyvrb}
\usepackage[dvipdfm]{graphicx}
\usepackage{amssymb}
\usepackage{endnotes}

\let\footnote=\endnote

\title{Neural realisation of the SP theory: cell assemblies revisited}

\author{J Gerard Wolff\\
\\
{\small \em CognitionResearch.org.uk,}\\
{\small \em E-mail: jgw@cognitionresearch.org.uk,}\\
{\small \em Telephone: +44(0)1248 712962.}}

\begin{document}

\maketitle

\begin{abstract}

This paper describes how the elements of the SP theory \citep{wolff_icmaus_overview} may be realised with neural structures and processes. To the extent that this is successful, the insights that have been achieved in the SP theory-the integration and simplification of a range of phenomena in perception and cognition-may be incorporated in a neural view of brain function. 

These proposals may be seen as a development of Hebb's \citeyearpar{hebb_1949} concept of a `cell assembly'. By contrast with that concept and variants of it, the version described in this paper proposes that any one neuron can belong in one assembly and only one assembly. A distinctive feature of the present proposals is that any neuron or cluster of neurons within a cell assembly may serve as a proxy or reference for another cell assembly or class of cell assemblies. This device provides solutions to many of the problems associated with cell assemblies, it allows information to be stored in a compressed form, and it provides a robust mechanism by which assemblies may be connected to form hierarchies, grammars and other kinds of knowledge structure.

Drawing on insights derived from the SP theory, the paper also describes how unsupervised learning may be achieved with neural structures and processes. This theory of learning overcomes weaknesses in the Hebbian concept of learning and it is, at the same time, compatible with the observations that Hebb's theory was designed to explain.
\end{abstract}

\section{Introduction}\label{introduction_section}

At its most abstract level, the SP theory \citep{wolff_icmaus_overview,wolff_2001_igpl,wolff_pothos_commentary} is intended to model {\em any} kind of system for processing information, either natural or artificial. The theory is Turing-equivalent in the sense that it can model the operation of a Universal Turing Machine \citep{wolff_1999_comp} but, unlike earlier theories of computing, the SP theory provides an account of a range of phenomena in perception and cognition, including the analysis and production of natural language \citep{wolff_2000}, `fuzzy' recognition of patterns and objects, probabilistic kinds of reasoning, solving problems by reasoning and by analogy \citep{wolff_1999_prob}, and unsupervised learning \citep{wolff_cavtat_2003}. The theory also provides a new perspective on a range of concepts in computing, logic and mathematics \citep{wolff_1999_comp, wolff_maths_logic}.

In work to date, the SP theory has been developed in purely abstract terms without reference to the anatomy or physiology of neural tissue. The main purpose of this paper is to consider possible ways in which the abstract concepts that have been developed within the SP theory may be mapped on to structures and mechanisms in the brain. To the extent that this is successful, we may achieve a `neural' version of the theory---called `SP-neural'---that inherits the insights that are provided by the abstract version.

To anticipate a little, it is proposed that `patterns' in the SP theory are realised with structures resembling Hebb's \citeyearpar{hebb_1949} concept of a `cell assembly'. By contrast with that concept:

\begin{itemize}

\item It is proposed that any one neuron can belong in one assembly and {\em only} one assembly. However, any given assembly may be `referenced' from other assemblies somewhat in the same way that any given web page may be referenced by URLs on other web pages.\footnote{In both cases, recursive self-referencing is also possible, as discussed in Section \ref{natural_language_section}.} This device provides solutions to many of the problems associated with cell assemblies, it allows information to be stored in a compressed form, and it provides a robust mechanism by which assemblies may be connected to form hierarchies, grammars and other kinds of knowledge structure.

\item The mechanisms for learning in SP-neural are significantly different from the well-known principles of Hebbian learning. However, they are compatible with the observations that Hebb's theory was designed to explain and, at the same time, they provide explanations for observations that the Hebbian theory is not able to explain.

\end{itemize}

The next section describes the SP theory in outline with just sufficient detail for present purposes. Section \ref{neural_realisation_section} considers how the elements of the SP theory may be realised with neural mechanisms. Section \ref{discussion_section} considers a selection of issues that arise in validating the proposals against empirical data and alternative theories. Section \ref{conclusion_section} summarises explanatory benefits, empirical predictions and future avenues for research.

\section{Outline of the SP theory}\label{sp_theory_section}

In this section, the main elements of the SP theory are described with many details omitted. The focus is on those aspects of the theory that are relevant to the proposals that follow. 

The SP theory is founded on two quasi-independent areas of thinking:

\begin{itemize}

\item A long tradition in psychology that many aspects of perception and cognition may be understood in terms of information compression \citep[see, for example,][]{attneave_1954, oldfield_1954, barlow_1959, barlow_1969, wolff_1988, chater_1996, chater_1999}.

\item Principles of {\em minimum length encoding}\footnote{An umbrella term for `minimum message length encoding' and `minimum description length encoding'.}, pioneered by \citet{solomonoff_1964, wallace_boulton_1968, rissanen_1978} and others, that focus on the intimate connection that exists between information compression and the inductive prediction of the future from the past \citep[see also][]{li_vitanyi_1997, solomonoff_1997}.

\end{itemize}

In broad terms, the SP theory is conceived as an abstract system or model that works like this. It receives `New' data from its environment and adds these data to a body of stored knowledge called `Old'. At the same time, it tries to compress the information as much as possible by searching for full or partial matches between patterns and merging or `unifying' patterns or parts of patterns that are the same. In the course of trying to compress information, the system builds {\em multiple alignments}, as described below.

Generally speaking, New information may be equated with sensory information but information within the system may sometimes play the same r{\^o}le, as described in Section \ref{production_section}, below.

\subsection{Computer models}

Two computer models of the SP system have been developed:

\begin{itemize}

\item SP61 is a partial realisation of the theory that builds multiple alignments and calculates probabilities of multiple alignments but does not transfer any information from New to Old: all its Old information must be supplied by the user. This model, which is relatively robust and stable, is described quite fully in \citet{wolff_2000}.

\item SP70 realises all the main elements of the theory, including the transfer of information from New to Old. This model, and its application to unsupervised learning, is described in \citet{wolff_cavtat_2003, wolff_unsupervised_learning}. More work is required to realise the full potential of this model.

\end{itemize}

\subsection{A `universal' format for knowledge}\label{universal_format_for_knowledge}

All information in the system is expressed as arrays or {\em patterns} of symbols, where a {\em symbol} is simply a `mark' that can be compared with any other symbol to decide whether it is the `same' or `different'. In work done to date, the focus has been on one-dimensional strings or sequences of symbols but it is envisaged that, at some stage, the ideas will be generalized to patterns in two dimensions.

This very simple `universal' format for knowledge has been adopted with the expectation that it would facilitate the representation of diverse kinds of knowledge and their seamless integration. Notwithstanding the simplicity of the format, the processing mechanisms provided within the SP system mean that it is possible to model such things as grammatical rules (with `context-sensitive' power), if-then rules, discrimination networks and trees, class-inclusion hierarchies (with inheritance of attributes), and part-whole hierarchies. An example will be seen in Section \ref{ma_section}, below, and others may be found in the sources cited above.

Another motive for adopting this format for knowledge is the observation that much of our knowledge derives ultimately from sensory inputs, especially vision, and most sensory inputs map naturally on to sequences or two-dimensional arrays. This idea sits comfortably with the observation that the cortex is, topologically, a two-dimensional structure and that there is a relatively direct mapping between the retina and each of the areas of the visual cortex, between the skin and the somatosensory cortex, and likewise for other senses. It seems reasonable to suppose that this kind of mapping is a general feature of the way the brain handles information.\footnote{No attempt will be made in this paper to consider at any length how we may encode our knowledge of the three-dimensional shapes of objects or the arrangement of objects in three-dimensional space. It is conceivable that 3D information might be encoded with 3D patterns but in biological terms, this may not make good sense. Given the predominantly two-dimensional nature of the retina and the cortex, it is more plausible to suppose that we encode our knowledge of three dimensional structures using two-dimensional patterns in the manner of plans and elevations used by architects and engineers.}

\subsection{Building multiple alignments}\label{ma_section}

The main elements of the multiple alignment concept as it has been developed in this research are illustrated in the example presented here. For the sake of clarity and to save space, this example is relatively simple. However, this should not be taken to represent the limits of what the system can do. More complex examples may be found in \citet{wolff_2000} and the other sources cited above.

In any realistic case, the number of possible alternative alignments is far too large to be searched exhaustively. It is necessary to use heuristic methods that prune away large parts of the search space, trading accuracy for speed. The SP61 and SP70 models both use forms of `hill climbing' with measures of compression to guide the search.

Given a New pattern representing the sentence `t h e c a t s l e e p s' and a set of Old patterns representing grammatical rules, the SP system builds multiple alignments like the one shown in Figure \ref{alignment_1}. The aim is to create multiple alignments that allow the New pattern to be encoded economically in terms of the Old patterns as described in Section \ref{evaluation_of_alignments}, below. Out of the several alignments that SP61 has built in this case, the one shown in Figure \ref{alignment_1} is the best. 

\begin{figure*}[!bhpt]
\fontsize{06.00pt}{07.20pt}
\begin{center}
\begin{BVerbatim}
0                        t h e                    c a t                        s l e e p   s   0
                         | | |                    | | |                        | | | | |   |  
1                        | | |                    | | |              < Vstem 2 s l e e p > |   1
                         | | |                    | | |              |   |               | |  
2                        | | |                    | | |       < V Vs < Vstem             > s > 2
                         | | |                    | | |       | | |                          |
3 S Num     ; < NP       | | |                    | | |     > < V |                          > 3
     |      | | |        | | |                    | | |     |     |                           
4    |      | | |        | | |          < Nstem 3 c a t >   |     |                            4
     |      | | |        | | |          |   |           |   |     |                           
5    |      | | |        | | |   < N Ns < Nstem         > > |     |                            5
     |      | | |        | | |   | | |                    | |     |                           
6    |      | | |  < D 0 t h e > | | |                    | |     |                            6
     |      | | |  | |         | | | |                    | |     |                           
7    |      | < NP < D         > < N |                    > >     |                            7
     |      |                        |                            |                           
8   Num SNG ;                        Ns                           Vs                           8
\end{BVerbatim}
\end{center}
\caption{The best alignment found by SP61 with `t h e c a t s l e e p s' in New and patterns representing grammatical rules in Old.}
\label{alignment_1}
\end{figure*}

By convention, the New pattern is always shown in row 0 of any alignment, as can be seen in Figure \ref{alignment_1}. The Old patterns are shown in the rows below the top row, one pattern per row. The order of the Old patterns across the rows is entirely arbitrary and without any special significance.

A pattern like `$<$ NP $<$ D $>$ $<$ N $>$ $>$' in row 7 of the alignment expresses the idea that a noun phrase (`NP') is composed of a determiner (`D') followed by a noun (`N'). In a context-free phrase-structure grammar, this would be expressed with a rule like `NP $\rightarrow$ D N'.

\begin{figure*}[!hbt]
\begin{center}
\includegraphics[width=10cm,height=6cm]{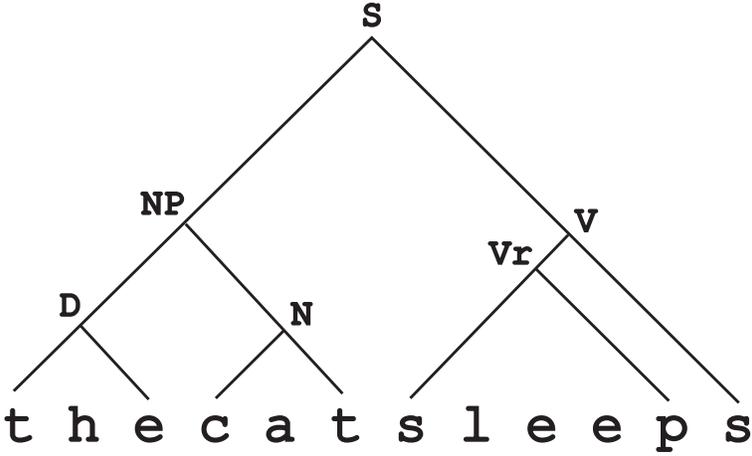}
\end{center}
\caption{A tree-structured parsing equivalent to the alignment shown in Figure \ref{alignment_1}, excluding row 8.}
\label{tree_parsing_figure}
\end{figure*}

If we ignore row 8, the whole alignment may be seen to achieve the effect of a context-free parsing, dividing the sentence into its constituent words, identifying `t h e c a t' as a noun phrase, marking each word with its grammatical class, and, within the verb `s l e e p s', marking the distinction between the stem and the suffix. Ignoring row 8, the alignment in Figure \ref{alignment_1} is equivalent to the tree-structured parsing shown in Figure \ref{tree_parsing_figure}.

\subsubsection{Context-sensitive power}\label{context_sensitive_section}

Although some of the patterns in the alignment are similar to rules in a context-free phrase-structure grammar, the whole system has the expressive power of a context-sensitive system. This is illustrated in row 8 of the alignment where the pattern `Num SNG ; Ns Vs' marks the `number' dependency between the singular noun in the subject of the sentence (`c a t') and the singular verb (`s l e e p s'). A basic context-free phrase-structure grammar, without augmentation, cannot handle this kind of dependency.

\subsubsection{`Identification' symbols and `contents' symbols}

Within each pattern in Old, there is a distinction between {\em identification} (ID) symbols and {\em contents} (C) symbols. The former serve to identify the pattern or otherwise define its relationship with other patterns, while the latter represent the contents or substance of the pattern.

For example, in the pattern `$<$ NP $<$ D $>$ $<$ N $>$ $>$' in row 7 of Figure \ref{alignment_1}, the ID-symbols are the initial left bracket (`$<$'), the symbol `NP' that follows it and the terminating right bracket (`$>$'). All other symbols in the pattern are C-symbols. In the pattern `$<$ Nstem 3 c a t $>$', the ID-symbols are the initial and terminal brackets, as before, plus the symbols `Nstem' and `3', while the C-symbols in that pattern are `c', `a' and `t'. In the pattern `Num SNG ; Ns Vs', the first three symbols are ID-symbols and the last two are C-symbols.

In SP61---which does not attempt any learning---the Old patterns are supplied to the system by the user and each symbol within each Old pattern is marked by the user to show whether it is an ID-symbol or a C-symbol. In SP70---which creates patterns and adds them to Old in the course of learning---the distinction between ID-symbols and C-symbols within any one pattern is marked by the system when the pattern is created.

\subsubsection{Evaluation of alignments}\label{evaluation_of_alignments}

A combination of ID-symbols, derived from an alignment like the one shown in Figure \ref{alignment_1}, can provide an abbreviated code for the entire sentence pattern (or other pattern) held in New. A `good' alignment is one where this code is relatively small in terms of the number of bits of information that it contains. The `compression score' for an alignment is the difference between the size (in bits) of the New pattern in its original form and its size after it has been encoded.

The procedure for deriving an encoding from an alignment is quite simple: scan the alignment from left to right looking for columns containing a single instance of an ID-symbol, not matched to any other symbol. The encoding is simply the symbols that have been found by this procedure, in the same order as they appear in the alignment. The encoding derived in this way from the alignment in Figure \ref{alignment_1} is `S SNG 0 3 2'. This is smaller than the New pattern in terms of the number of symbols it contains. It is even smaller when the size of the New pattern and the size of the code are measured in terms of the numbers of bits they contain, as described in \citet{wolff_2000}.

The `compression score' or `compression difference' for an alignment is:
\begin{equation}
CD = N_o - N_e
\end{equation}
where $N_o$ is the size (in bits) of the New pattern in its original form and $N_e$ is its size in bits after it has been encoded.

\subsection{Production of sentences and other patterns}\label{production_section}

An attractive feature of the SP system is that, {\em without any modification}, it can support the production of language (or other patterns of knowledge) as well as its analysis. If SP61 is run again, with the sentence in New replaced by the encoded form of the sentence (`S SNG 0 3 2') as described in Section \ref{evaluation_of_alignments}, the best alignment found by the system is exactly the same as before except that row 0 contains the encoded pattern and each symbol in that pattern aligned with matching symbols in the rows below. The original sentence has, in effect, been recreated because the alignment contains the words of the sentence in their correct order. This is an example of the possibility noted near the beginning of Section \ref{sp_theory_section}) where the r{\^o}le of `New' information is played by information within the system rather than by sensory data.

It is envisaged that the production of sentences from meanings may be modelled in a similar way.  Instead of using a code pattern like `S SNG 0 3 2' to drive the production process, some kind of semantic structure may be used instead.

\subsection{Unsupervised learning}\label{sp70_section}

The SP system as a whole is a system that learns by assimilating `raw' information from its environment and distilling the essence of that information by a process of information compression. The main elements of the theory are now realised in the SP70 computer model. This model is able to abstract simple grammars from appropriate data without any kind of external `teacher' or the provision of `negative' samples or the grading of samples from simple to complex \citep[cf.][]{gold_1967}. In short, it is an {\em unsupervised} model of learning. Some reorganisation is needed to overcome certain weaknesses in the model as it stands now.

Although some reorganisation is needed to overcome certain weaknesses in SP70 as it stands now, the overall structure appears to be sound. In the model, learning occurs in two stages: 

\begin{enumerate}

\item {\em Creation of Old patterns}. As the system receives New patterns, a variety of Old patterns are derived from them as explained below. Some of these patterns are `good' in terms of principles of minimum length encoding but many of them are `bad'.

\item {\em Selection of `good' patterns}. By a process of sifting and sorting through the Old patterns, the system abstracts one or more subsets, each one of which is relatively good in terms of the principles of minimum length encoding. The patterns in the first one or two of these subsets may be retained and the remaining patterns may be discarded.

\end{enumerate}

It is envisaged that, in future versions of the model, these two processes---creation of Old patterns and selection amongst them---will be repeated many times while New patterns are being received so that the system can gradually bootstrap a set of Old patterns that are relatively useful for the economical encoding of New data.

To get the flavour of the way in which Old patterns are created, consider a simple example. If the current pattern in New is `t h e b o y r u n s' and the repository of Old patterns is empty, the system discovers that there is no way to encode the New pattern economically in terms of Old information so it augments the New pattern with ID symbols (which converts it into `$<$ \%1 t h e b o y r u n s $>$') and adds the augmented pattern to Old. When Old has accumulated a range of patterns like this, it can begin to create multiple alignments. If, at that stage, the current pattern in New is `t h e g i r l r u n s', multiple alignments created by the system will include one like this:

\begin{center}
\begin{BVerbatim}
0      t h e g i r l r u n s   0
       | | |         | | | |  
1 < %1 t h e b o y   r u n s > 1
\end{BVerbatim}
\end{center}

\noindent From this alignment, the system can derive some additional Old patterns by extracting coherent sequences of matched symbols and unmatched symbols, adding system-generated ID-symbols, and creating another pattern that ties everything together. 

In this case, the result is five patterns like this: 

\begin{center}
\begin{BVerbatim}
< %2 t h e >
< %3 r u n s >
< %4 0 b o y >
< %4 1 g i r l >
< %5 < %2 > < %4 > < %3 > >
\end{BVerbatim}
\end{center}

\noindent The first four of these patterns represent words while the last one is an `abstract' pattern that describes the overall structure of two original sentences in terms of ID-symbols in the lower-level patterns.

Notice how `b o y' and `g i r l' have been assigned to the same class (identified by the symbol `\%4'), very much in the tradition of distributional linguistics \citep[see, for example,][]{harris_1951, fries_1952}.

Notice also how the system picks out entities that we naturally regard as `discrete' \citep[see also][]{wolff_1977, wolff_1980, brent_1999}. Similar principles may explain how we learn to see the visual world in terms of discrete objects.

The flexible pattern matching that is built into the process of finding multiple alignments means that SP70 can find correlations that bridge arbitrary amounts of intervening structure. It is not necessary to restrict the learning process to second- or third-order correlations ({\em cf.} \citet{thompson_2000}).

\subsection{Other aspects of perception and cognition}\label{other_aspects_section}

Apart from the parsing and production of language and the unsupervised learning of knowledge, the SP system has proved to be a versatile model of several other aspects of perception and cognition:

\begin{itemize}

\item {\em Fuzzy pattern recognition and best-match retrieval of information}. The system provides a model for the way we can recognise things despite errors of omission, commission and substitution and the way we can retrieve information from fragmentary clues.

\item {\em Recognition at multiple levels of abstraction}. Given appropriate patterns in its store of Old information, the system can be used to model the recognition of things at several different levels of abstraction simultaneously. Some unknown entity may be recognised as an individual, `Tibs', and at the same time it may be recognised as a cat, a mammal, a vertebrate and an animal (see also the example described at the end of Section \ref{sharing_section}).

\item {\em Probabilistic and exact forms of reasoning}. The system supports probabilistic `deduction', abduction, nonmonotonic reasoning and chains of reasoning. It also provides a model for classical `exact' forms of reasoning.

\item {\em Solving problems by reasoning and by analogy}. Given appropriate patterns, the system can do such things as finding a route between two places or solving problems by analogy.

\end{itemize}

Further information may be found in \citet{wolff_icmaus_overview} and earlier publications cited there.

\subsection{`Identification' and `Reference' in the representation and processing of knowledge}\label{id_ref_section}

The SP system expresses a pair of ideas that are fundamental in the representation and processing of knowledge and are one of the corner-stones of the proposals in this paper:

\begin{itemize}

\item {\em Identification}. Any entity or concept may be given a `name', `label', `code', `tag' or {\em identifier} by which it may be identified. Examples in everyday life include the name of a person, town, country, book, theorem, poem, type of animal or plant, period of history, and many more.

The concept of identification is not restricted to unique identification of specific things. It also applies to the identification of {\em classes} of things like `furniture', `people', `animals' and so on

\item {\em Reference}. Whenever we wish to refer to any such entity or class, we can do so by means of a {\em copy} of the relevant name, label or identifier. Every copy is a {\em reference} to the given entity or class and in each case there may be many such references. Examples in everyday life include referring to someone by their name, or referring to a town, country, book etc by its name. In a paper like this one, 
`\citet{hebb_1949}' is a reference to the bibliographic details at the end of the paper and those details are themselves a reference to the book itself.

\end{itemize}

Being able to refer to any concept by a relatively brief name is a great aid to succinct communication. Imagine how cumbersome and difficult things would be if we had to give a full description of everything and anything whenever we wanted to talk about it---like the slow language of the Ents in Tolkien's {\em The Lord of the Rings}. Identification and reference are a powerful aid to the compression of information and it should not surprise us to find that the same pair of ideas lies at the heart of computer-based `ZIP' programs for compressing information.

In the SP system, any given pattern has an identifier in the form of one or more ID-symbols. One or more copies of those symbols in other patterns or in the same pattern serve as references to that pattern. In Figure \ref{alignment_1}, the identifier for the pattern `$<$ NP $<$ D $>$ $<$ N $>$ $>$' in row 7 are the symbols `$<$' and `NP' at the beginning of the pattern and the symbol `$>$' at the end. A reference to that pattern appears as `$<$ NP $>$' in the pattern `S Num ; $<$ NP $>$ $<$ V $>$' in row 3. In a similar way, `$<$ \%2 $>$' and `$<$ \%3 $>$' in the pattern `$<$ \%5 $<$ \%2 $>$ $<$ \%4 $>$ $<$ \%3 $>$ $>$' from Section \ref{sp70_section} may be seen as references to `$<$ \%2 t h e $>$' and `$<$ \%3 r u n s $>$', respectively, and `$<$ \%4 $>$' is a reference to the class \{`$<$ \%4 0 b o y $>$',`$<$ \%4 1 g i r l $>$'\}.

\section{Neural realisation of the SP concepts}\label{neural_realisation_section}

This section describes how the SP concepts may be realised in terms of neural structures and processes. Much of the discussion in this section relates to particular domains---mainly vision and language---but it should be emphasised that the proposals are intended to apply to sensory data and knowledge of all kinds---visual, auditory, tactile etc---both individually and in concert.

In this section and the rest of the paper, the focus will be on the analysis of sensory data rather than the production of sentences or other patterns of knowledge. In every figure, arrows on neural connections will show the direction of flow of sensory signals. A full discussion of the way the system may model the creation or production of motor patterns is outside the scope of this paper. In brief, it is envisaged that codes or `meanings' may take on the r{\^o}le of New information as described in Section \ref{production_section} and they may serve to create neural analogues of multiple alignments by precisely the same processes that are described in Section \ref{creating_neural_mas}, below. 

\subsection{New patterns and symbols}\label{new_patterns_and_symbols}

Apart from the kinds of `internal' data just mentioned, New information in the SP system corresponds to sensory data. The discussion in this subsection is mainly about vision but similar principles seem to apply to other sensory modalities. 

In mammalian vision, patterns of light entering the eye are received by the retina and transmitted via the lateral geniculate body to layers 3 and 6 of the visual cortex and beyond. Although sensory information is received initially in analogue form, it is widely accepted that the information is converted at an early stage into something like David Marr's \citeyearpar{marr_1982} concept of a `primal sketch'. In the early stages, sensory data is converted into the digital language of on-centre cells, off-centre cells and the like, and in later stages it is interpreted in terms of digital features such as lines at particular angles, motion in a particular direction, and so on \citep[see, for example,][]{nicholls_etal_2001, hubel_2000}.

Each neuron in the cortex that responds to a particular feature of the sensory data may be regarded as the neural equivalent of an SP symbol and we shall call it a {\em neural symbol}. An array of neural symbols that registers a sensory pattern in any sensory cortex we shall call a {\em receptor array}. Whatever its three dimensional configuration, we shall assume that it is topologically equivalent to a one-dimensional sequence or a two-dimensional array. A receptor array equates with the part of the SP system that receives New patterns from the system's environment and passes it on for processing within the system. 

\begin{figure*}[!hbt]
\centering
\includegraphics[width=11cm,height=5cm]{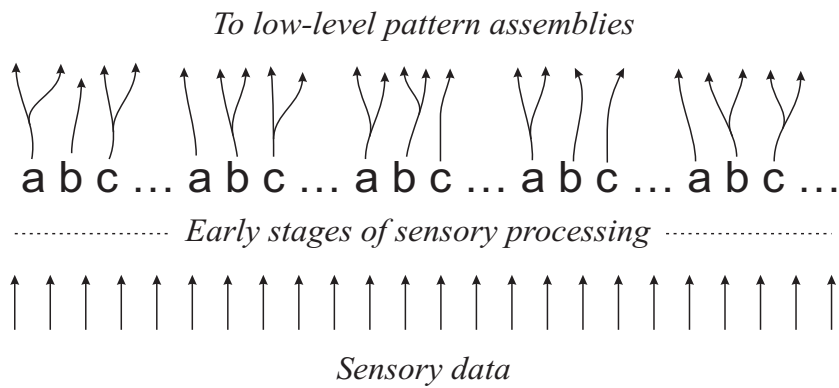}
\caption{Schematic representation of the repeating alphabet of neural symbols in a receptor array (shown as repetitions of `a b c ...'). Also shown is the path taken by sensory data through low-level sensory processing to the receptor array and onwards to low-level pattern assemblies.}
\label{receptor_array_figure}
\end{figure*}

Notice that, if a receptor array is to respond to a wide variety of patterns---which is clearly true of the visual cortex and other sensory cortices---then the complete `alphabet' of neural symbols must be available at every location within the array and thus repeated many times across the receptor array as shown schematically in Figure \ref{receptor_array_figure}. In accordance with this expectation, it has been found that---with the type of neuron that responds selectively to a short line segment at a particular angle---the complete range of orientations is repeated across the cortex within each of a large number of fairly small `orientation columns' \citep{barlow_1982, hubel_2000}. It seems likely that a similar organisation may exist in the somatosensory cortex, mirroring the way in which receptors that respond selectively to heat, cold, touch, pressure and so on are repeated across areas of the skin \citep[][Chapter 18]{nicholls_etal_2001}.

So far, we have assumed that a New pattern arrives all at once like an image projected on to the retina. But it is clear that streams of sensory information are being received in all sensory modalities throughout our lives and that we are very sensitive to patterns in the temporal dimension within those streams of information, especially in hearing. It is assumed here that temporal sequences in hearing, vision and other modalities is captured spatially in arrays of neurons, probably in the cortex.

\subsection{Old patterns and symbols}\label{old_patterns_and_symbols}

An Old pattern in the SP system may be realised by a network of interconnected cells in the cortex, similar to Donald Hebb's \citeyearpar{hebb_1949} concept of a {\em cell assembly} \citep[see also][pp. 103--107]{hebb_1958}. Because there are significant differences between the present proposals and Hebb's original concept, the neural realisation of an Old pattern will be referred to as a {\em pattern assembly}. Those differences are described in this subsection and ones that follow, and they are further discussed in Section \ref{discussion_section}.

An SP {\em symbol} in an Old pattern may be realised within a pattern assembly by a single neuron or, perhaps, a small network of interconnected neurons. To simplify discussion in the rest of this paper, we shall assume that every symbol is realised by a single neuron and, as before, we may refer to any such neuron as a {\em neural symbol}. A neuron within a pattern assembly that corresponds to a C-symbol within an SP pattern will be called a `C neural symbol' or, more simply, a {\em C-neuron}. Likewise, a neuron that represents an ID-symbol will be called an `ID neural symbol' or an {\em ID-neuron}.

\sloppy As with receptor arrays, it is assumed that, regardless of the three-dimensional configuration of the neurons in pattern assemblies, each such assembly is topologically equivalent to a sequence of neurons or a two-dimensional array of neurons.

Unlike the Hebbian concept of a cell assembly, it is envisaged that the neurons in any one pattern assembly lie close together within the cortex. It is envisaged that the connections between neurons will be two-way excitatory connections mainly between immediate neighbours but there may also be excitatory connections between neural symbols that are a little more distant. In general, all the interconnections amongst neurons in a pattern assembly will be relatively short. However, there will be connections between pattern assemblies as described in Section \ref{connections_section} and those connections may be very much longer.

The suggested configuration of pattern assemblies just described is consistent with Braitenberg's \citeyearpar{braitenberg_1978} suggestion (pp. 181-182) that there is strong correlation of neural activity within each 1 mm column in the human cortex and that each column may be connected to any other column. We may suppose that each pattern assembly is confined within one column but connections between pattern assemblies may span the whole cortex.

As with SP patterns, pattern assemblies may be roughly graded from `low level' to `high level'. Low level pattern assemblies are those representing small perceptual details such as formant transitions and allophone in the case of speech or small perceptual `motifs' such as corners of objects, textures, or colour combinations in the case of vision. High level pattern assemblies would represent abstractions like the `sentence' pattern in row 3 of Figure \ref{alignment_1} or high-level classes like `the world', `people' or `the animal kingdom'. As with SP patterns, the gradation is not a strict hierarchy because any given pattern assembly may have a direct connection to any other pattern assembly, not necessarily ones that are immediately `above' or `below'.

\subsection{Connections}\label{connections_section}

It is envisaged that each neural symbol within a receptor array sends signals to one or more C-neurons in one or more pattern assemblies. Most of these would be pattern assemblies that may be classified as low level but in principle any pattern assembly may receive signals from a receptor array. Connections leaving the neural symbols of a receptor array are shown at the top of Figure \ref{receptor_array_figure}. 

\begin{figure*}[!hbt]
\centering
\includegraphics[width=11cm,height=15cm]{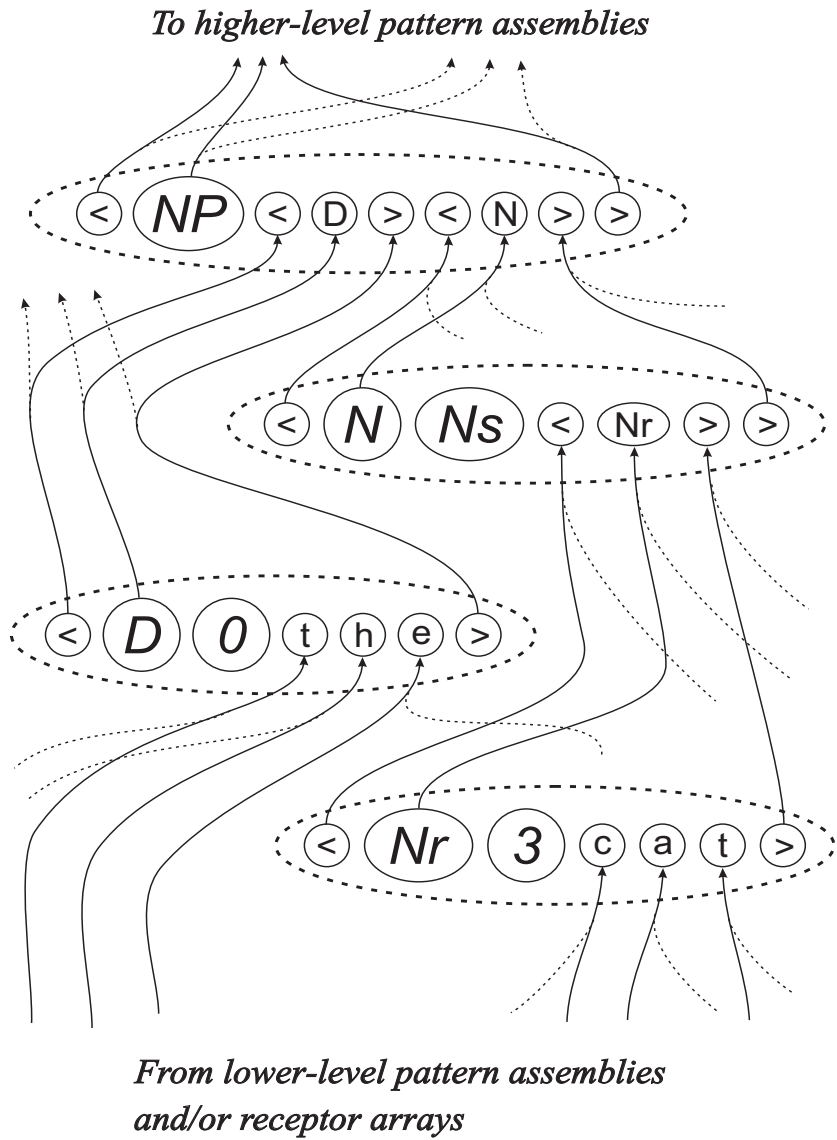}
\caption{Schematic representation of inter-connections amongst pattern assemblies. Envelopes with broken lines are pattern assemblies and envelopes with unbroken lines are neurons. Arrows show the direction of flow of sensory signals. ID-neurons (apart from those representing brackets) are larger than C-neurons and contain oblique characters. Not shown in the figure are lateral connections within each pattern assembly and inhibitory connections as described in Section \ref{creating_neural_mas}.}
\label{connections_figure}
\end{figure*}

Each C-neuron within each pattern assembly may receive signals from one or more receptor arrays or from one or more pattern assemblies or from both kinds of structure. Normally, these signals would come from lower level structures but in principle any one pattern assembly may receive signals from any of the entire set of pattern assemblies, including itself. Each ID-neuron within each pattern assembly may send signals to one or more C-neurons. Normally, these signals would go to higher level structures but in principle any one pattern assembly may send signals to any of the entire set of pattern assemblies, including itself. These kinds of interconnections amongst pattern assemblies are shown schematically in Figure \ref{connections_figure}, following the conventions shown in the legend to the figure. 

An aspect of the proposals that is not illustrated in the figure is that there will be low level pattern assemblies representing structures in every sensory modality---vision, hearing, touch etc---and high-level pattern assemblies may receive inputs from any of these lower level pattern assemblies. In this way, any high-level concept like `person' may be described in terms of all the relevant sensory modalities. In many cases, the connections that are needed to bring diverse sensory modalities together in this way will be relatively long, as previously indicated. 

An important feature of the connections between a receptor array and a pattern assembly and the connections between pattern assemblies is that {\em the symbols at each end of every connection are always a match for each other}. In effect, the r{\^o}le of each C-neuron is {\em defined} by its inputs. For example, the neural symbol `a' in the pattern assembly `$<$ Nstem 3 c a t $>$' in Figure \ref{connections_figure} may be seen to represent `a' because it receives inputs from other neural symbols that represent `a'. Because the r{\^o}le of each C-neuron within each pattern assembly is defined in this way, it is not necessary to provide a repeated alphabet of symbols at each location within each pattern assembly like the repeated alphabet of symbols that is necessary at each location of a receptor array (Section \ref{new_patterns_and_symbols}). In a similar way, the r{\^o}le of each ID-neuron is defined by the pattern assemblies to which it connects.

The fact that connections between two pattern assemblies are always between symbols that match each other means that the proposed system of inter-connections amongst pattern assemblies provides a neural realisation of the relationship between identification and reference, as described in Section \ref{id_ref_section}. For example, the symbols `$<$ D $>$' in the pattern `$<$ NP $<$ D $>$ $<$ N $>$ $>$' in Figure \ref{connections_figure} are connected to the matching symbols in the pattern `$<$ D 0 t h e $>$' and it is envisaged that there would be similar connections to other patterns representing words in the grammatical class of `determiners'. Thus `$<$ D $>$', as a reference to that grammatical class, has neural connections to patterns that represent the class.

It will be convenient to use the term {\em neural reference} for one or more neurons that is the neural equivalent of a reference, as described in Section \ref{id_ref_section}.

\subsection{Creating neural analogues of multiple alignments}\label{creating_neural_mas}

Given a structure of receptor arrays, pattern assemblies and their interconnections as described so far, how can it function to identify the neural equivalent of one or two `good' multiple alignments like the one shown in Figure \ref{alignment_1}?

It is envisaged that each pattern assembly will respond to incoming stimulation in a manner that is functionally coherent and similar to the responses of an individual neuron. Below a certain threshold, there will be relatively little response amongst neurons in the pattern assembly. But when inputs exceed the threshold, the pattern assembly will `fire' and all the neurons in the assembly will become active together (`ignition' as described by \citet{pulvermuller_book_2002}). 

It is envisaged that the neural equivalent of a multiple alignment will be identified when each of its pattern assemblies and the connections between them are relatively active as suggested by the connections shown with unbroken lines in Figure \ref{connections_figure}. Other pattern assemblies and other connections will be relatively quiet, as suggested by the connections shown with broken lines in the same figure.

It seems possible that there may also be a system of inhibitory connections amongst pattern assemblies---perhaps mediated by structures below the cortex---that would dampen down excessive neural activity \citep{milner_1957} and would have the effect of forcing a decision amongst competing alternatives so that the `winner takes all'. A mechanism of that sort operates in the innervation of the Basilar membrane, where position along the length of the membrane encodes the frequency of sound and lateral inhibition has the effect of strengthening the response where the signal is strongest whilst suppressing the response in neighbouring regions \citep{von_bekesy_1967}.

Another mechanism that may help to prevent excessive excitation of pattern assemblies are inhibitory connections amongst the input fibres to pattern assemblies, as described in Section \ref{order_relations_perception}, next.

\subsubsection{Keeping track of order in perception}\label{order_relations_perception}

As the system has been described so far, a stimulus like `b o y' would produce the same level of response in a pattern assembly like `$<$ N 2 y o b $>$' as it would in a pattern assembly like `$<$ N 3 b o y $>$'. In other words, there is nothing in what has been proposed to keep track of the ordering or arrangement of elements in a pattern.

Our ability to recognise patterns despite some jumbling of their elements (e.g., solving anagrams) is consistent with the scheme that has been described. But the fact that we can see the difference between an anagram and its solution shows that something else is needed. Some kind of mechanism is required to distinguish between two kinds of situation:

\begin{enumerate}

\item[(a)] Fibres leave a pattern assembly or one portion of a receptor array and arrive together, in the same relative positions, at one portion of another pattern assembly. Fibres that conform to this rule will be described as {\em coherent}.

\item[(b)] Fibres arrive at one portion of a pattern assembly from a variety of different sources or, if they come from one source, their relative positions are not preserved. This kind of arrangement will be described as {\em incoherent}.

\end{enumerate}

When signals arrive at a pattern assembly, they should produce a stronger response in the first case than in the second. A possible mechanism to ensure that this happens would be lateral connections amongst the fibres of a coherent bundle that would have the effect of increasing the rate of firing in the bundle if the majority of them are firing together. When any one fibre in the bundle is firing, then signals will be carried to neighbouring fibres via the lateral connections and these signals should lower the threshold for firing and thus increase the rate of firing in any of the neighbouring fibres that are already firing. An alternative scheme that should achieve the same effect is inhibitory connections amongst fibres that are incoherent.

If signals arrive in a temporal sequence, then other possible mechanisms include `synfire chains' and temporal `sequence detectors', as described by \citet{pulvermuller_2002, pulvermuller_book_2002}. However, these appear to be less suitable for keeping track of order in spatially-distributed visual patterns, seen in a single glance.

\subsection{Learning}\label{neural_learning_section}

It is envisaged that learning in SP-neural would occur in a manner that is similar to SP70 (Section \ref{sp70_section}):

\begin{itemize}

\item When there is a good match between sensory patterns and stored pattern assemblies, the sensory data will be encoded in terms of those pattern assemblies much as in Section \ref{evaluation_of_alignments}. But if there is no good match for sensory data or only a partial match, then new pattern assemblies may be created like the patterns described in Section \ref{sp70_section}.

\item The creation of a new pattern assembly is unlikely to mean growing new neurons from scratch. Although there is evidence that new nerve cells can grow in mature brains \citep{shors_gould_2001}, it seems likely that the bulk of neurons in the brain are present at birth and that the creation of a new pattern assembly is largely a process of assigning pre-existing neurons to it, with the creation or breaking of connections as outlined below. 

\item Associated with each pattern assembly will be some physiological analogue of the frequency measure associated with each SP pattern---counting the number of times each pattern assembly has been recognised in sensory inputs. Pattern assemblies that have been frequently recognised should have a lower threshold for excitation or should be otherwise more responsive to incoming stimulation.

\item Periodically, pattern assemblies may be evaluated in terms of their usefulness for encoding sensory data. Those that are proving useful may be retained and those that are not proving useful may be purged. The purging of a pattern assembly would not mean the literal destruction of the neurons in the assembly. It would be largely a matter of breaking the connections with other pattern assemblies so that the neurons in the pattern assembly can be made available for later re-use in some other pattern assembly.

\end{itemize}

With regard to the making and breaking of connections, the lateral connections amongst the neural symbols of a new pattern assembly may already be in place before it is created and the new assembly may be brought into existence by the relatively simple process of breaking connections at the boundaries. Something like this is suggested by evidence for the progressive weakening of connections between cells that do not normally fire at the same time \citep[][pp. 254--255]{pulvermuller_1999} except that, in the present proposal, connections at the boundaries would be broken completely when a pattern assembly is first created. We shall return to this and related points in Section \ref{learning_discussion}.
 
Each new pattern assembly would also need appropriate connections with receptor arrays and other pattern assemblies. At first sight, this suggests the growing of new fibres, many of which would be quite long. It is much more plausible to suppose, as Hebb suggested, that there are pre-established long connections between different parts of the cortex and that new connections are established by the growth of short connecting links to the main connections (much as a new telephone may normally be installed without the necessity of laying a new cable all the way to the exchange).

It may not even be necessary to postulate the growth of short links. It is possible that pre-existing links could be switched on or off somehow according to need. Something like this might enable us to account for the speed with which we can establish memories for names, faces, events and so on (see Section \ref{learning_discussion}, below).

\section{Discussion and evaluation}\label{discussion_section}

This section considers a selection of issues in the evaluation of the SP-neural proposals as they relate to empirical phenomena and alternative proposals, mainly Hebb's \citeyearpar{hebb_1949} original concept of a `cell assembly' and versions of that concept that have been proposed subsequently. To avoid terminological confusion, I will use the term `cell assembly' for all versions of that concept (including pattern assemblies) unless a particular version is the focus of interest.

\subsection{Parts, wholes, classes and associations}\label{sharing_section}

Hebb \citeyearpar{hebb_1949} proposed that cell assemblies could become associated and that assemblies that are associated might eventually merge to become a single assembly. He also envisaged an hierarchical organisation for conceptual structures and the possibility that low-level concepts (and their corresponding cell assemblies) might be shared by two or more higher-level concepts (and their corresponding cell assemblies) \citep[see also][pp. 103--107]{hebb_1958}.

In this connection, \citet{milner_1996} raises some pertinent questions: 

\begin{quote}

``How do associations between cell assemblies differ from internal associations that are responsible for the assemblies' properties? It does not seem likely that both these processes can be the result of similar synaptic changes as is usually assumed. If they were, the interassembly associations would soon become intraassembly loops. A related puzzle is that parts are not submerged in the whole. Doors and windows are integral parts of my concept of a house, but they are also robust, stand-alone concepts.'' (p. 71).

\end{quote}

\noindent Later on the same page he writes: ``Perhaps the toughest problem of all concerns the fact that we have many associations with almost every engram....The brain must be a veritable rat's nest of tangled associations, yet for the most part we navigate through it with ease.''

\citet{hebb_1958} provides a possible answer to the way in which parts may be distinguished from wholes:

\begin{quote}

``If two assemblies A and B are repeatedly active at the same time they will tend to become `associated,' so that A excites B and vice versa. If they are always active at the same time they will tend to merge in a single systems---that is, form a single assembly---but if they are also active at different times they will remain separate (but associated) systems. (This means that exciting part of A, for example, has a very high probability of exciting all of A, but a definitely lower probability of exciting a separate assembly, B; A may be able to excite B only when some other assembly, C, also facilitates activity in B).'' (p. 105).

\end{quote}

\begin{figure*}[!hbt]
\centering
\includegraphics[width=10cm,height=15cm]{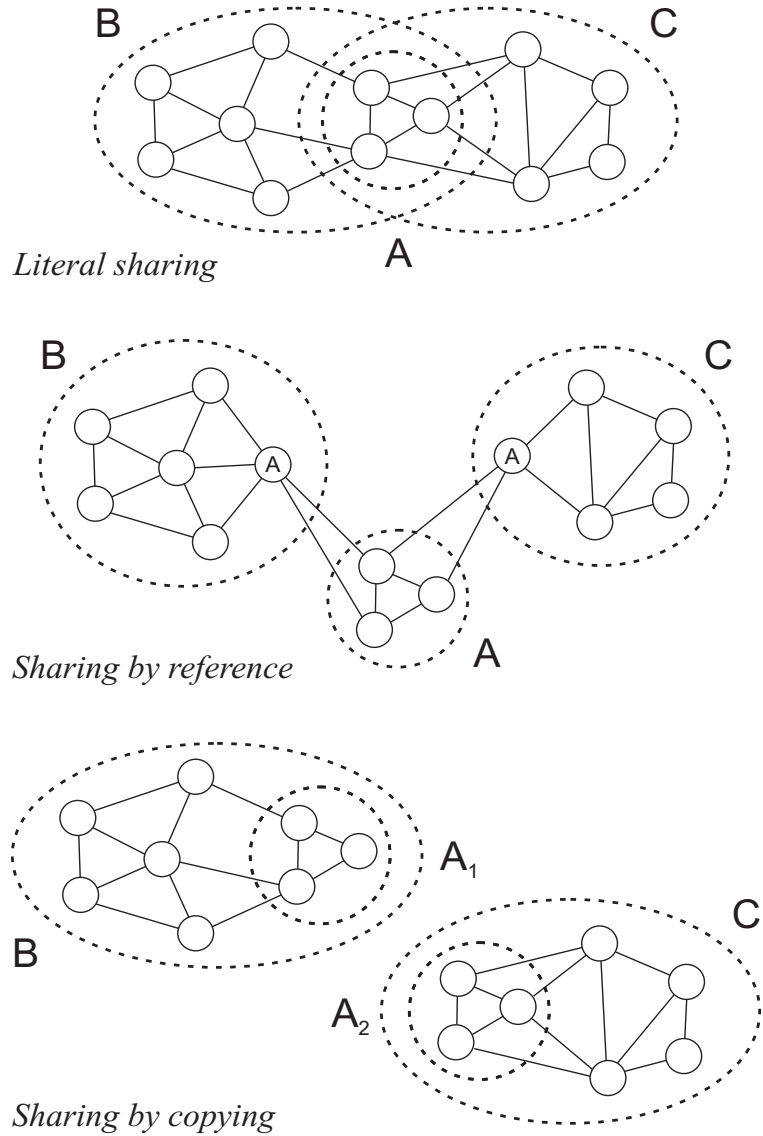}
\caption{Three possible ways in which a low-level cell assembly (`A') may be shared by two higher-level cell assemblies (`B' and `C'). Cell assemblies are enclosed with broken-line envelopes and neurons are shown with unbroken lines.}
\label{sharing_figure}
\end{figure*}

\noindent This may be part of the answer but it does not get to the bottom of the problem. There is a need to recognise that sharing of structures can be done in three distinct ways as illustrated in Figure \ref{sharing_figure} and described here:

\begin{itemize}

\item {\em Literal sharing}. The neurons in the lower-level cell assembly (`A' in the figure) are also part of two or more higher-level cell assemblies (`B' and `C' in the figure).

\item {\em Sharing by reference}. The lower-level cell assembly (`A') is {\em outside} the higher-level cell assemblies (`B' and `C') but each of the latter contains one or more neurons (shown in the figure as a single neuron marked with a small `A') that serves as a proxy, agent or representative for the lower-level cell assembly and is connected to that assembly. The proxy is a neural reference, as described in Section \ref{connections_section}---functionally equivalent to the concept of a reference described in Section \ref{id_ref_section}.

If sharing of structures is always achieved in this way---as proposed in this paper---then any one neuron belongs in one cell assembly and {\em only} one cell assembly.

\item {\em Sharing by copying}. In each of two or more higher-level cell assemblies (`B' and `C') there is a {\em copy} of the lower-level cell assembly (`A'). As with sharing by reference, there is no need to postulate that any one neuron may belong in two or more cell assemblies.

\end{itemize}

It is reasonably clear that Hebb and all subsequent authors have not intended the third sense in which cell assemblies might share structure. Although it may have a r{\^o}le in some aspects of cognition, I shall say no more about it in this paper.

With regard to the first two senses in which cell assemblies may share structure, they have not to my knowledge been clearly differentiated either in Hebb's writings or in any other writings about cell assemblies. Hebb's descriptions of the cell assembly concept imply that any one neuron may belong in two or more cell assemblies---in accordance with the first sense of structure sharing. This seems to be generally accepted by people writing later and it has often been made explicit (see, for example, \citet[p. 216]{palm_1982}, \citet[p. 385]{huyck_2001a}, \citet[pp. 257--258]{pulvermuller_1999}, \citet[p. 213]{sakurai_1998} and \citet{gardner-medwin_barlow_2001}).

But as a means of representing conceptual structures and their inter-relations, literal sharing is problematic. For example, it is difficult to see why assemblies B and C in the top part of Figure \ref{sharing_figure} should not simply merge into a single cell assembly. Perhaps more serious is the problem of keeping track of the relative positions or ordering of shared components in two or more configurations. 

\begin{figure*}[!hbt]
\centering
\includegraphics[width=10cm,height=12cm]{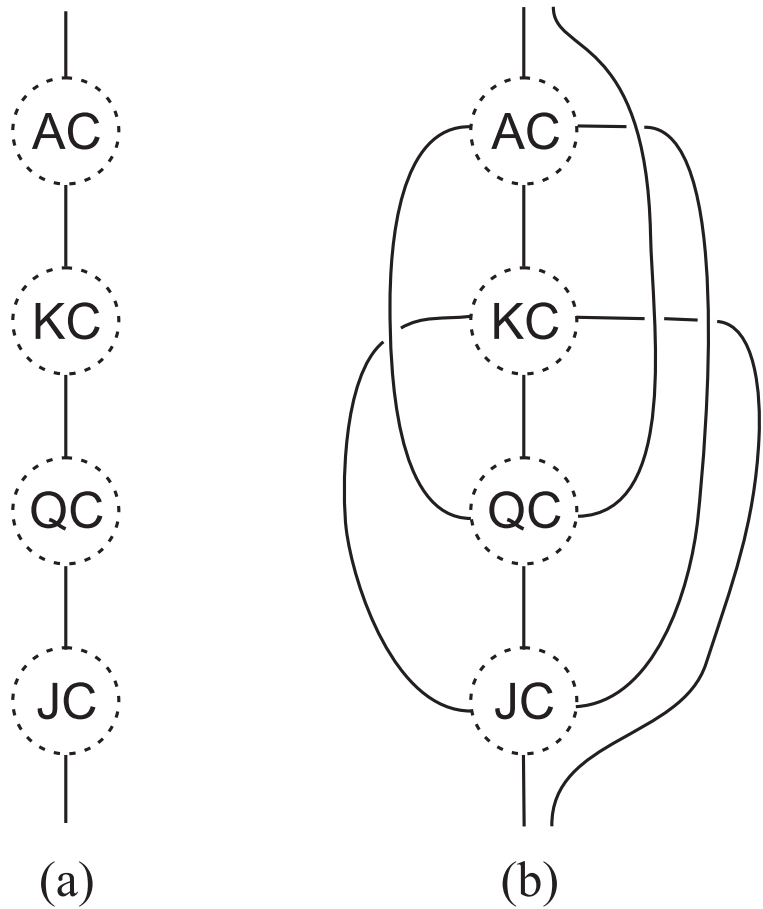}
\caption{{\em Representing two orderings using literal sharing}. (a) Part of a cell assembly representing four playing cards and their order within a deck of cards before it has been shuffled. (b) As in (a) but with the addition of connections showing a new order for the same four cards after the pack has been shuffled. {\em Key}: `A' = Ace, `K' = King, `Q' = Queen, `J' = Jack, `C' = Clubs. Each playing card is itself represented by a cell assembly shown as a broken-line circle.}
\label{cards_sharing_literal_figure}
\end{figure*}

Consider a pack of playing cards containing the Ace, King, Queen and Jack of Clubs, all together and in that order. This configuration may be represented within a cell assembly as shown in part (a) of Figure \ref{cards_sharing_literal_figure}. Here, each card is represented by a cell assembly and the order of the cards within the pack is shown by connections between those cell assemblies.

As it stands, the representation is unambiguous. But we run into difficulties if, using literal sharing of structures, we try to add a representation of the pack after it has been shuffled. Part (b) of Figure \ref{cards_sharing_literal_figure} shows the result if the same four cards fall together in the pack but in the order Queen, Ace, Jack and King. The addition of connections to represent that order causes the whole structure to become ambiguous. If the connections are non-directional, the Ace, for example, may be followed immediately by any of the other three cards. And even if directional links are allowed, the Ace may be followed immediately by the King or the Jack.

\begin{figure*}[!hbt]
\centering
\includegraphics[width=10cm,height=12cm]{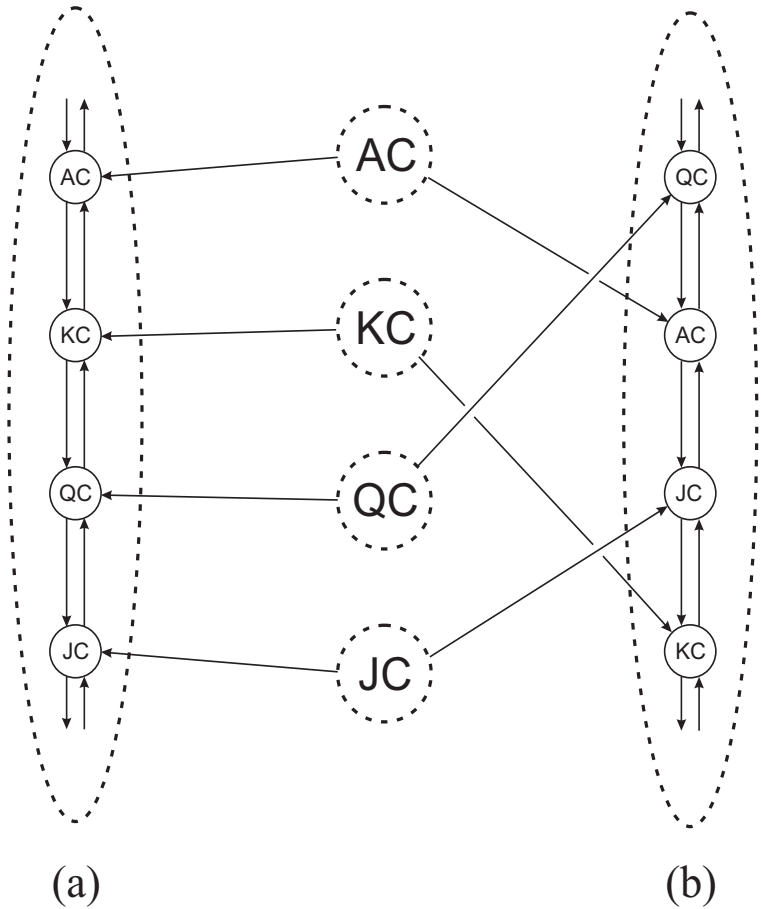}
\caption{{\em Representing two orderings using sharing by reference}. (a) Part of a cell assembly representing a pack of playing cards before shuffling. Four C-neurons are shown, each of which serves as a neural reference to a cell assembly in the middle of the figure representing one playing card. (b) Part of a cell assembly representing the same pack of cards after shuffling. As before, four C-neurons are shown and each one serves as a neural reference to one of the cell assemblies in the middle of the figure. The abbreviations are the same as in Figure \ref{cards_sharing_literal_figure}.}
\label{cards_sharing_reference_figure}
\end{figure*}

The problem can be overcome if, instead of using literal sharing of structures, we use sharing by reference. Parts (a) and (b) of Figure \ref{cards_sharing_reference_figure} each show how an ordering of the playing cards may be represented unambiguously by a sequence of neurons, each one of which is a neural reference to a cell assembly for the corresponding card (shown in the middle of the figure).\footnote{I am grateful to Daniel Wolff for suggesting the pack-of-cards example.}

Neural references also provide a neat solution to the other problems raised by Milner:

\begin{itemize}

\item Associations between pattern assemblies may be encoded by building a new pattern assembly containing neural references to the pattern assemblies that are to be associated. These associations are then internal to the new pattern assembly and the original pattern assemblies retain their identity. 

\item In a similar way, there can be a stand-alone cell assembly for each component of a house while the assembly for the house itself comprises a collection of neural references to the components. In this way, the concept of a house does not become muddled with concepts for doors, windows etc.

\item It is true that there must be a `rat's nest' of associations in the brain but neural references allow these to be encoded in a way that does not disturb the integrity of each concept. The coherence of a web page is not affected by links to that page from other pages and it makes no difference how many such pointers or references there may be.

\end{itemize}

\begin{figure*}[!hbt]
\centering
\includegraphics[width=10cm,height=11cm]{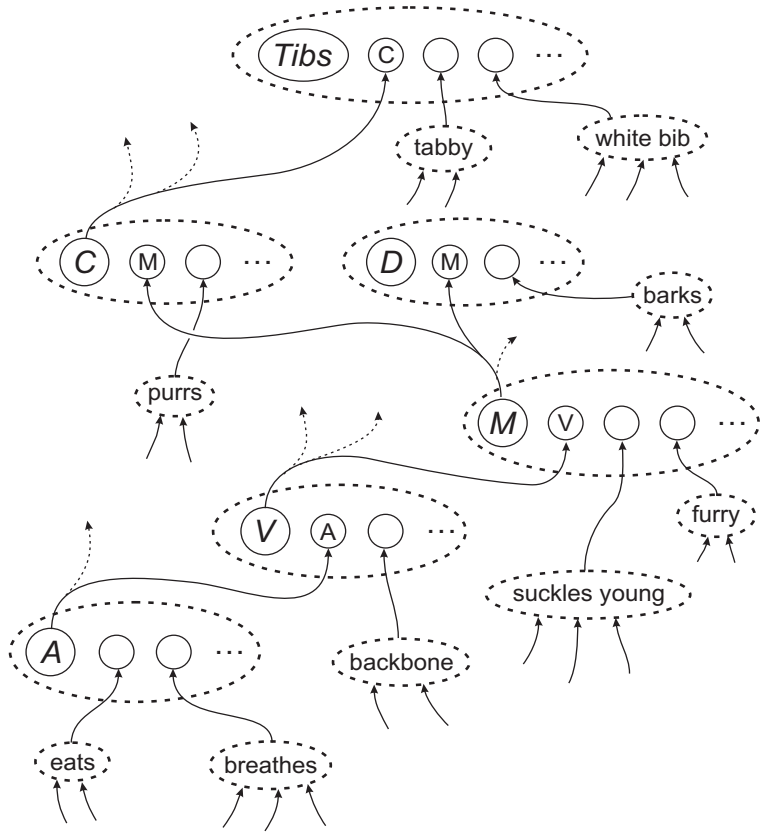}
\caption{An example showing how pattern assemblies with neural references between them can represent class-inclusion relations and part-whole relations. {\em Key}: `C' = cat, `D' = dog, `M' = mammal, `V' = vertebrate, `A' = animal, `...' = further structure that would be shown in a more comprehensive example. The conventions are the same as in Figure \ref{connections_figure} except that brackets have been omitted to simplify the figure and neurons are not shown in the lower-level pattern assemblies.}
\label{class_part_figure}
\end{figure*}

Some of the power of sharing by reference can be seen in Figure \ref{class_part_figure}. The figure shows a hierarchy of classes from the relatively abstract `animal' (`A' near the bottom of the figure), via `vertebrate' (`V'), `mammal' (`M'), `cat' (`C'), to a specific cat `Tibs' shown near the top of the figure. At the same time, the figure shows the part-whole relations between each class and the descriptive elements of the class. In a more elaborate example, these elements may themselves be broken down into a hierarchy of parts and subparts. As previously noted, arrows on the connections between pattern assemblies show the direction of flow of sensory signals.

In the figure, each C-neuron in the body of a pattern assembly functions as a proxy for or neural reference to the pattern assembly from which it receives sensory signals. The whole structure achieves the effect of `inheritance' in an object-oriented system: the individual cat called `Tibs' inherits the feature `purrs' from the `cat' pattern assembly, the features `suckles young' and `furry' are inherited from the `mammal' pattern assembly, `backbone' is inherited from the `vertebrate' assembly, and `eats' and `breathes' are inherited from the `animal' pattern assembly.
\subsubsection{Temporal correlations?}
An alternative to the foregoing is to suppose that cell assemblies may form hierarchies and other kinds of structure by temporal correlations amongst their neurons: ``A set of units can be bound into a block by synchronising their fast activity fluctuations. Several such blocks can coexist if their activity is desynchronised relative to each other ...'' \citep[][p. 15]{von_der_malsburg_1987} and ``To represent that Attribute A is bound to Attribute B and Attribute C to Attribute D, the cells for A and B fire in synchrony, the cells for C and D fire in synchrony, and the AB set fires out of synchrony with the CD set.'' \citep[][p. 485]{hummel_bierdeman_1992}.

Ideas of this kind---reviewed by \citet{bienenstock_geman_1995}---are clearly very different from the SP-neural proposals but there is insufficient space here for any kind of weighing of the pros and cons.
\subsection{Cardinal cells}
ID-neurons in the present proposals are similar to the `cardinal cells' described by \citet{barlow_1972}, except that there may be more than one for each concept. Constructs such as neural reference and multiple alignment are also distinctive in the present proposals.
The chief objection to the idea of cardinal cells is that, if `grandmother' is represented by a single cell, we will not be able to access our knowledge of her if that cell dies. But as Barlow points out, a small amount of replication will give considerable protection against this kind of catastrophe. As suggested in Section \ref{old_patterns_and_symbols}, each neural symbol may be a small cluster of neurons rather than a single neuron. Another possibility is that all our concepts are replicated two or three times in the brain. Arguments and calculations that would take too much space to reproduce here suggest that, with the SP-neural scheme for representing knowledge, this kind of replication is feasible.

\subsection{Learning}\label{learning_discussion}

Hebb's proposals for learning are summarised in his much-quoted suggestion that:

\begin{quote}

``When an axon of cell A is near enough to excite a cell B and repeatedly or persistently takes part in firing it, some growth process or metabolic change takes place in one or both cells such that A's efficiency, as one of the cells firing B, is increased.'' \citep[][p. 62]{hebb_1949}.

\end{quote}

\noindent This idea forms the basis for learning in many varieties of artificial neural network. Apart from the elegant simplicity of the idea, another possible reason for its popularity is that it accords with statistical models of learning where two things need to co-occur with a relatively high frequency before the association between them can be seen to be significant. And the Hebbian view of learning is broadly in accordance with the slow build up of our knowledge throughout childhood and beyond.

And yet this view of learning is in direct conflict with everyday observations of what we can and do learn from a single occurrence or experience. If we are involved in some kind of accident which does not cause us to lose consciousness, we can remember the sequence of events and recount them with little difficulty. A possible objection here is that, where strong emotions are involved, we may rehearse our memories many times and thus strengthen the links amongst the corresponding cell assemblies. But the same objection carries much less force if we are asked to recall things that carry less emotional charge. On any given evening, we normally have no difficulty in recalling the humdrum events of the day and this without any apparent rehearsal.

Because the slow growth of cell assemblies does not account for our ability to remember things immediately after a single exposure, Hebb adopted a `reverberatory' theory for this kind of memory. But, as \citet{milner_1996} points out, it is difficult to understand how this kind of mechanism could explain our ability to assimilate a previously-unseen telephone number. Each digit in the number may be stored in a reverberatory assembly but this does not explain how we remember the {\em sequence} of digits in the number.

In the learning scheme outlined in Sections \ref{sp70_section} and \ref{neural_learning_section}, new pattern assemblies can be created in response to a single sensory input. This is consistent with our ability to remember unique events and sequences of events. And, if we suppose that synapses can be switched on or off in seconds or fractions of a second according to need, we may be able to account for the speed with which immediate memories can be established. This model for short-term memory does not suffer from the weakness in the reverberatory model that Milner identified: we may remember the sequence of digits in a telephone number by creating a new pattern assembly that represents the number.

But an ability to lay down new pattern assemblies relatively quickly does not explain why it takes several years to learn something like a language. The principles on which SP70 is based suggest why learning a language, and similar kinds of learning, take quite a lot of time. The abstract `space' of alternative grammars for any natural language is astronomically large and, even using heuristic techniques like those in the SP models, it takes time to search amongst the many possibilities. Finding a tolerably good grammar for any natural language is a very complex problem and it cannot be solved in an instant.

\subsection{Constancies}\label{constancies}

A possible objection to the SP-neural proposals is that they are inconsistent with the `constancy' phenomena in perception. These include:

\begin{itemize}

\item {\em Size constancy}. We can recognise an object despite wide variations in the size of its image on the retina---and we judge its size to be constant despite these variations.

\item {\em Brightness constancy}. We can recognise something despite wide variations in the absolute brightness of the image on our retina (and, likewise, we judge its intrinsic brightness to be constant).

\item {\em Colour constancy}. In recognising the intrinsic colour of an object, we can make allowances for wide variations in the colour of the light which falls on the object and the consequent effect on the colour of the light that leaves the object and enters our eyes.

\end{itemize}

If the pattern recorded in a receptor array was merely a copy of sensory input there would indeed be wide variations in the size of the visual pattern projected by a given object and similar variations in brightness and colours. The suggestion here is that much of the variability of sensory input from a given object has been eliminated at a stage before the information reaches the receptor array. Lateral inhibition in the retina emphasises boundaries between relatively uniform areas. The redundant information within each uniform area is largely eliminated which means that it is, in effect, shrunk to the minimum size needed to record the attributes of that area. Since this minimum will be the same regardless of the size of the original image, the overall effect should be to reduce or eliminate variations in the sizes of images from a given object.

In a similar way, the `primal sketch' created by lateral inhibition should be largely independent of the absolute brightness of the original image---because it is a distillation of {\em changes} within an image, independent of absolute values. And adaptation in the early stages of visual processing should to a large extent prevent variations in brightness having an effect on patterns reaching the receptor array.

Edwin Land's `retinex' theory suggests that colour constancy is achieved by processing in the retina and in the visual cortex \citep[see][pp. 437--439]{nicholls_etal_2001}. This is consistent with the idea that this source of variability has been removed at a stage before sensory input is compared with stored patterns.

\subsection{Stimulus equivalence and discrimination}\label{stim_equiv_discrim}

An important motivation for the cell assembly concept was to explain the phenomenon of `stimulus equivalence' or `generalization'---our ability to recognise things despite variations in size, shape, position of the image on the retina, and so on. However, \citet{milner_1996} writes that ``Generalization is certainly an important and puzzling phenomenon, but discrimination is equally important and tends to be ignored in theories of neural representation. Any theory of engram formation must take into account the relationship between category and instance---the ability we have to distinguish our own hat, house, and dog from hats, houses and dogs in general.'' (p. 70).

Both stimulus equivalence and discrimination can be explained in terms of SP-neural. The theory accounts for generalization in four main ways:

\begin{itemize}

\item The way in which perceptual constancies may be accommodated in the present proposals was considered in Section \ref{constancies}. 

\item As with the original cell assembly concept, a pattern assembly can respond provided a reasonably large subset of its neurons has received inputs. This seems to accommodate our ability to recognise things despite omissions, additions or substitutions in sensory inputs.

\item In the present scheme, each neuron or feature detector in a receptor array is connected to each one of the pattern assemblies (mainly low level assemblies) that contain the corresponding feature. Thus, each of these pattern assemblies may respond to appropriate input regardless of the position of the input on the receptor array.

\item Owing to the provision of neural references in the present proposals, it is possible to create pattern assemblies that represent abstractions from sensory input. An example from Figure \ref{connections_figure} is the pattern assembly `$<$ NP $<$ D $>$ $<$ N $>$ $>$'. This pattern assembly represents the abstract structure of a noun phrase via references to the class of determiners (`$<$ D $>$') and the class of nouns (`$<$ N $>$'). It may be activated by a range of {\em alternative} inputs: any noun phrase that conforms to the determiner-noun pattern.

\end{itemize}

The SP theory provides a neat account of our ability to distinguish specific instances from the categories in which they belong. As mentioned in Section \ref{other_aspects_section}, the SP system can build multiple alignments which include several different levels of abstraction, including a level that represents a specific instance. Examples may be seen in \citet{wolff_icmaus_overview} and earlier publications. The way in which a hierarchy of classes (including a specific instance) may be represented in SP-neural is illustrated in the example described at the end of Section \ref{sharing_section} (Figure \ref{class_part_figure}).

\subsection{Natural language}\label{natural_language_section}

One of the strengths of the SP system is its ability to represent syntactic structures of natural languages and to process them in the parsing and production of sentences \citep{wolff_2000, wolff_icmaus_overview}. Although SP-neural is not complete, it already suggests solutions to the challenges for any neural theory of language processing that have been identified by \citet[pp. 144--146]{pulvermuller_book_2002}:

\begin{enumerate}

\item[(a)] How can centre-embedded sequences such as `ABCC$^\prime$B$^\prime$A$^\prime$' be represented?\footnote{Center embedding seems to be a genuine phenomenon in language even though most people can only cope with one or two levels of embedding.}

\item[(b)] How can discontinuous constituents and distributed words be realised (e.g., {\em switch ... on})?

\item[(c)] How is it possible to specify the syntactic relatedness of distant elements in a string (e.g., noun and verb agreement)?

\item[(d)] How can repeated use of the same word or lexical category within a sentence be modelled and stored? A related question is how recursive structures might be modelled and stored?

\item[(e)] How can lexical categories---such as `noun', `verb' and `adjective'---be realised in a neuronal network?

\end{enumerate}

The answers suggested here have similarities to and differences from those proposed by Pulverm{\"u}ller in chapters 8 to 12 ({\em ibid.}). There is insufficient space here for a detailed comparison so I will merely outline the way in which these questions may be answered in SP-neural.

The concept of neural reference provides a key to all five questions. SP-neural suggests that questions (a), (b), (d) and (e) may be answered in a way that is a close analogue of the organisation of a context-free phrase-structure grammar, while the `context sensitive' feature of the SP theory (Section \ref{context_sensitive_section}) may provide an answer to question (c).

\begin{figure*}[!hbt]
\centering
\includegraphics[width=8cm,height=12cm]{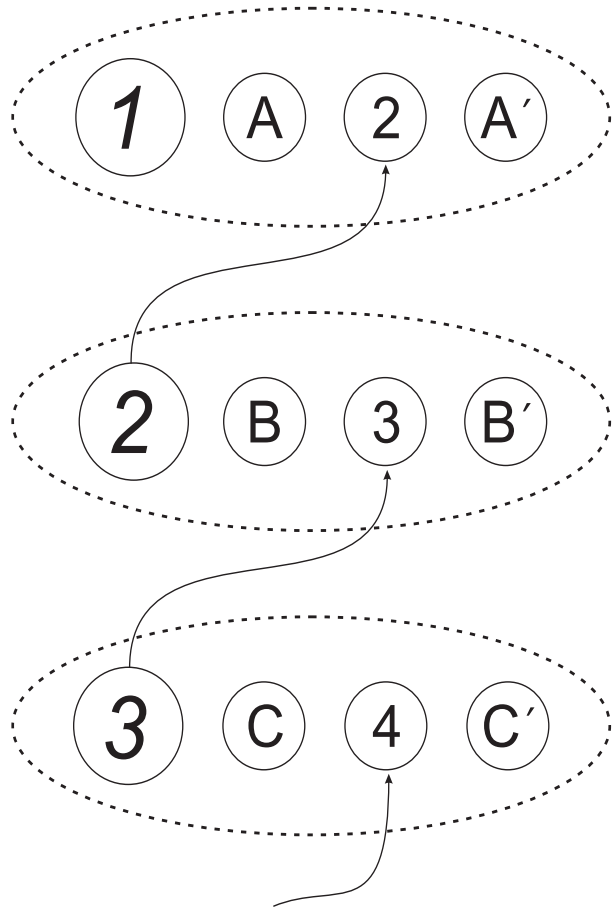}
\caption{Pattern assemblies and connections between them showing how centre-embedding may be modelled in SP-neural. In this and the remaining figures, the conventions are the same as in Figure \ref{class_part_figure}.}
\label{center_embedding_figure}
\end{figure*}

Figure \ref{center_embedding_figure} shows how,  using neural references, `B ... B$^\prime$' may be embedded in `A ... A$^\prime$' and how `C ... C$^\prime$' may be embedded in `B ... B$^\prime$'.

\begin{figure*}[!hbt]
\centering
\includegraphics[width=10cm,height=8cm]{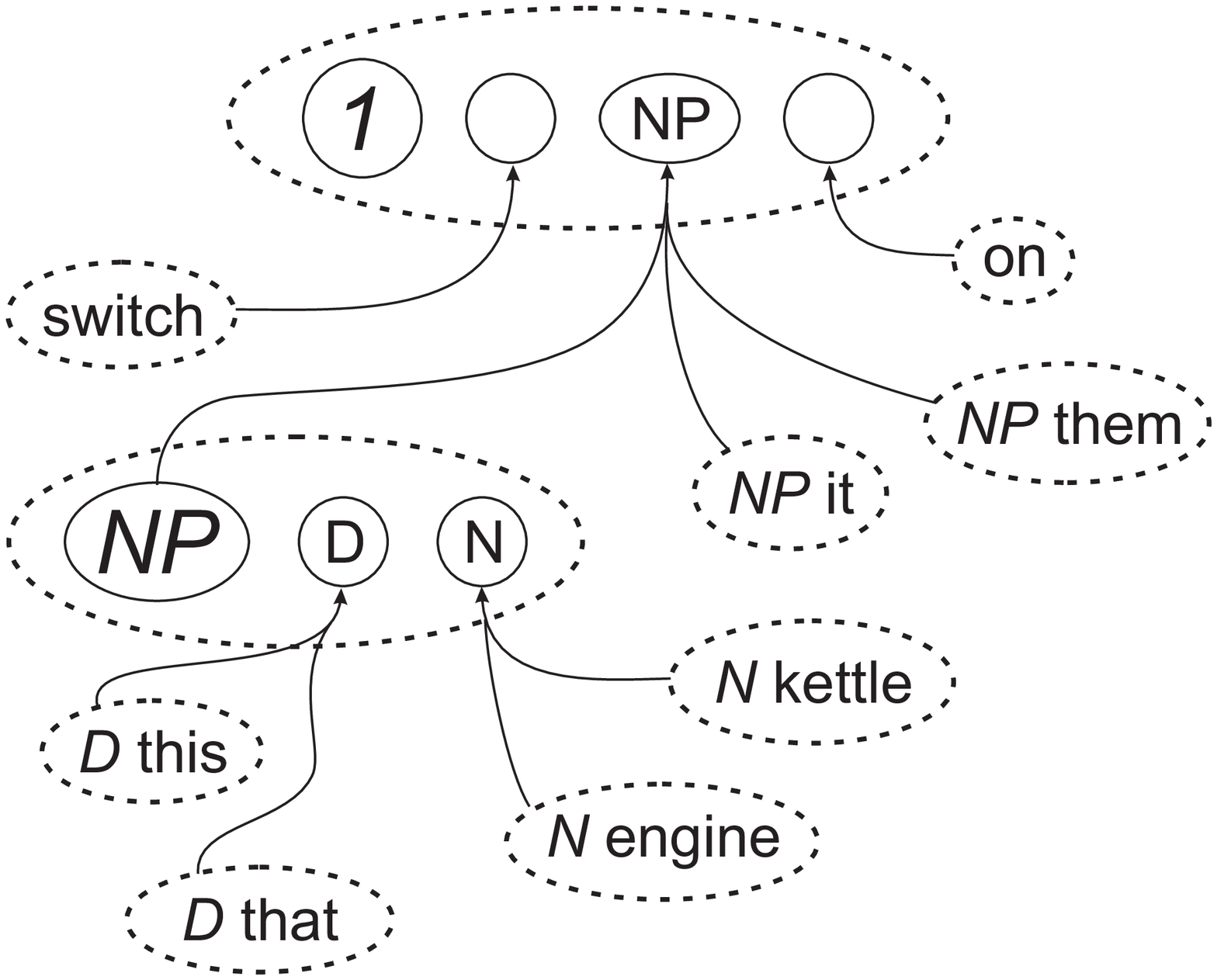}
\caption{Pattern assemblies and connections between them showing how discontinuous structures may be modelled in SP-neural.}
\label{discontinous_structures_figure}
\end{figure*}

In a similar way, the `embedding' of a noun-phrase pattern assembly within an assembly representing the phrase {\em switch ... on} may be achieved by means of a neural reference to the noun phrase placed between {\em switch} and {\em on}, as shown in Figure \ref{discontinous_structures_figure}. 

This figure also shows how grammatical categories such as `determiner' (`D') and `noun' (`N') may be represented in SP-neural. The pattern assembly representing the structure of a noun phrase contains neural references to each of these categories and each neural reference is connected to pattern assemblies representing the members of the category.

\begin{figure*}[!hbt]
\centering
\includegraphics[width=10cm,height=8cm]{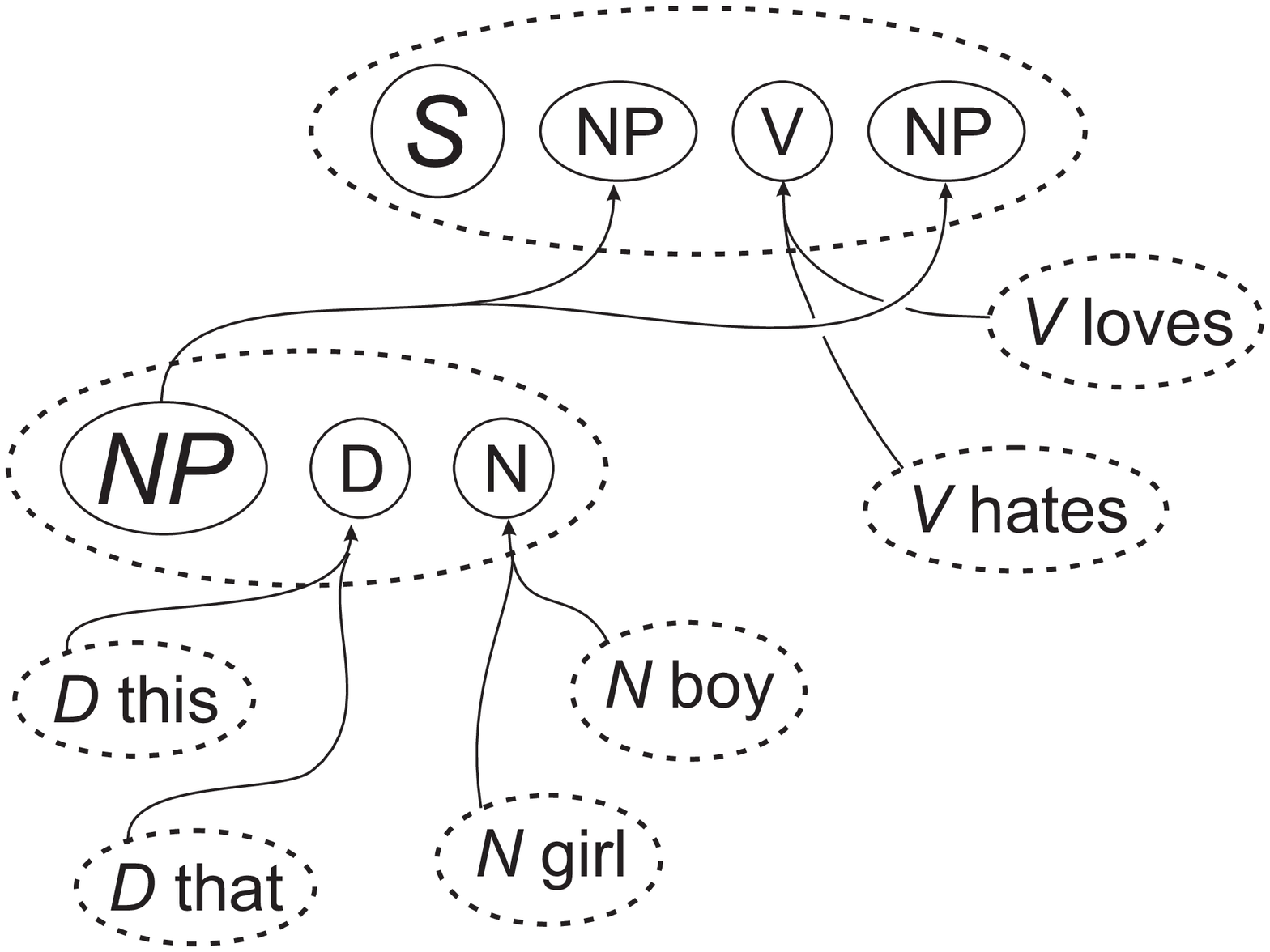}
\caption{Pattern assemblies and connections between them showing how repeated structures may be modelled in SP-neural.}
\label{repeated_structures_figure}
\end{figure*}

Figure \ref{repeated_structures_figure} shows how a noun phrase (`NP') may be repeated within a sentence by the provision of {\em two} neural references to a pattern assembly representing the structure of a noun phrase. The repetition is expressed by the neural references, not by the structure itself.

\begin{figure*}[!hbt]
\centering
\includegraphics[width=10cm,height=7cm]{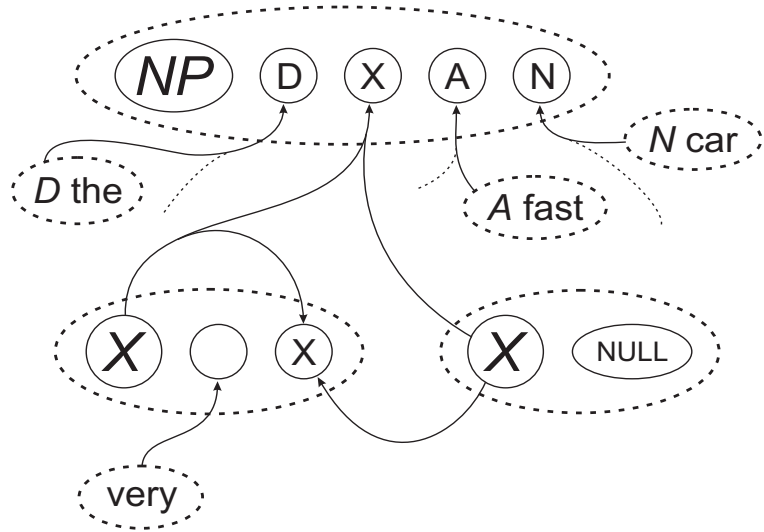}
\caption{Pattern assemblies and connections between them showing how recursion may be modelled in SP-neural.}
\label{recursion_figure}
\end{figure*}

And Figure \ref{recursion_figure} shows how the recursion in phrases like {\em the very very ... fast car} may be represented in SP-neural. The first of the two pattern assemblies identified as `X' contains a neural reference to itself and this expresses the recursive nature of `very ...'. Notice that, for the production of language but perhaps not for the analysis of language, it is also necessary to provide a `null' member of the `X' category---so that it is possible to escape from the recursive loop. These pattern assemblies are close analogues of the rules in a phrase-structure grammar that would be needed to describe this kind of language structure.

The suggested answer to question (c) within SP-neural hinges on the ability of the SP system to represent and process long-distance dependencies as indicated in Section \ref{context_sensitive_section} \citep[see also][]{wolff_2000, wolff_icmaus_overview}. It is envisaged that pattern assemblies representing grammatical structures may include ones like the pattern shown in row 8 of Figure \ref{alignment_1}. This pattern expresses the idea that, within a sentence, a singular noun in the `subject' position must be followed by a singular verb in the position of the main verb. There will also be a similar pattern representing plural dependencies. It is envisaged that, within SP-neural, these pattern assemblies will exert their influence in much the same way as the corresponding patterns in the SP system.

\section{Conclusion}\label{conclusion_section}

The broad outlines of SP-neural are reasonably clear but there are still several areas of uncertainty in the details. If this neural version of the SP theory can be completed satisfactorily, then it will inherit all the considerable explanatory force of the SP theory in several areas of perception and cognition including the way diverse forms of knowledge may be represented and integrated, the analysis and production of language, fuzzy recognition of objects, recognition at multiple levels of abstraction, retrieval of memories from fragmentary clues, various forms of probabilistic reasoning and exact reasoning, unsupervised learning, and the solving of problems by reasoning and by analogy.

Even though SP-neural is not complete, it already offers useful insights into the kinds of structures and functions that we may expect to find in brains and nervous systems. The theory makes a number of predictions that are, in principle, falsifiable although they represent a considerable challenge for current techniques and technologies. The main predictions of the theory are these:

\begin{itemize}

\item As in Hebb's theory and other theories in that tradition, SP-neural proposes that knowledge is stored in functionally-coherent and inter-connected assemblies of neurons in the cortex. However, the theory is distinctive in most of the points that follow.

\item It is proposed that any one neuron belongs in one pattern assembly and {\em only} one assembly.

\item It is envisaged that, while there may be long connections {\em between} pattern assemblies, the neurons within any one assembly lie close together within the cortex, perhaps confined to one column, with short connections between them, mainly between immediate neighbours. These are not a necessary part of the proposals but they seem likely.

\item Relationships amongst pattern assemblies, including part-whole relations, class-inclusion relations and associations between pattern assemblies, are mediated by the use of neurons that are functionally equivalent to {\em references} between structures, as described in the text.

\item When sensory information is received, recognition of patterns and objects is achieved by the excitation of pattern assemblies and connections between them in a manner that is functionally equivalent to the creation of multiple alignments in the SP theory.

\item There may be inhibitory connections between assemblies to dampen excessive excitation and sharpen competition between assemblies. Inhibitory connections between afferent fibres may help to preserve information about the order of features or events in sensory inputs.

\item It is proposed that learning is achieved by the creation of pattern assemblies and selection amongst them, in the same way that patterns are created and selected in the SP theory. Assemblies can be created and destroyed by the making and breaking of short connections, as described in text.

\item To record short-term memories, pattern assemblies may be created within seconds or fractions of a second, perhaps by the activation and de-activation of pre-existing synaptic connections. 

\end{itemize}

Finding direct neurophysiological evidence that may confirm of confute these predictions is not likely to be easy but the proposals in this paper may suggest new avenues for investigation. Uncertainties in the theory need to be resolved, ideally by observations of brain structure and functioning. Meanwhile, computer simulation techniques may help to shape the theory. 

\theendnotes

\section*{Acknowledgements}

I am very grateful for constructive comments on earlier versions of this paper 
that I have received from Horace Barlow, Jim Maxwell Legg, Peter Milner, Friedemann Pulverm{\"u}ller and Guillaume Thierry. The responsibility for all errors and oversights is, of course, my own. I am also grateful to Daniel Wolff for a very useful discussion of the sharing problem.

\raggedright

\end{document}